# Reasoning about the Value of Decision-Model Refinement: Methods and Application

**Kim Leng Poh***
Laboratory for Intelligent Systems
Department of Engineering-Economic Systems
Stanford University, CA 94305-4025

**Eric J. Horvitz**
Palo Alto Laboratory
Rockwell International Science Center
444 High Street, Palo Alto, CA 94301

## Abstract

We investigate the value of extending the completeness of a decision model along different dimensions of refinement. Specifically, we analyze the expected value of quantitative, conceptual, and structural refinement of decision models. We illustrate the key dimensions of refinement with examples. The analyses of value of model refinement can be used to focus the attention of an analyst or an automated reasoning system on extensions of a decision model associated with the greatest expected value.

## 1 Introduction

The quality of recommendations for action generated by decision analyses hinges on the fidelity of decision models. Indeed, the task of *framing* a decision problem—enumerating feasible actions, outcomes, uncertainties, and preferences—lies at the heart of decision analysis. Decision models that are too small or coarse may be blind to details that may have significant effects on a decision recommendation. Unfortunately, the refinement of decision models can take a great amount of time, and can be quite costly in time and expense. In some cases, actions are taken well before a natural stopping point is reached in the modeling process. In other cases, important distinctions about actions and outcomes are recognized days or months after a model is developed.

We have developed methods for probing the value of key dimensions of decision-model refinement. We pose the techniques as tools that can direct the attention of an analyst or of an automated reasoning system to refine aspects of a decision model along dimensions that have the highest expected payoff. The methods also can provide guidance on when it is best to cease additional refinement and to take immediate action in the world. Our work differs from previous studies of the value of modeling (Watson & Brown, 1978; Nickerson & Boyd, 1980) in that we develop a unifying framework for probing the values of different classes of refinement, and consider issues surrounding the direction of model building and improvement under resource constraints.

Three fundamental dimensions of decision-model refinement are (1) quantitative refinement, (2) conceptual refinement, and (3) structural refinement. We will explore methods for making decisions about which dimensions to refine, and the amount of effort to expend on each form of refinement.

*Quantitative refinement* is the allocation of effort to refine the uncertainties and utilities in a decision model. There are two classes of quantitative refinement: (1) uncertainty refinement, and (2) preference refinement. *Uncertainty refinement* is effort to increase the accuracy of probabilities in a decision model. For example, assessment may be focused on the tightening of bounds or second-order probabilities over probabilities in a decision model. *Preference refinement* is refinement of numerical values representing the utilities associated with different outcomes. For example, an analyst may work to refine his uncertainty about the value that a decision maker will associate with an outcome that has not been experienced by his client.

*Conceptual refinement* is the refinement of the semantic content of one or more distinctions in a decision model. With conceptual refinement, we seek to modify the precision or detail with which actions, outcomes, and related random variables are defined. For example, for a decision maker deliberating about whether to locate a party inside his home versus outside on the patio, it may be important to extend the distinction "rain" to capture qualitatively different types of precipitation, using such conceptually distinct notions as "drizzle," "intermittent showers," and "downpour." Likewise, with additional deliberation, he may realize that there are additional options available to him. Many of these additional alternatives are those that would not be taken if there were no uncertainty about the weather. For example, he might consider having the party on the porch, or renting a tent to shelter the guests in his yard.

---

*Currently at the Department of Industrial & Systems Engineering, National University of Singapore.

*Structural refinement* is modeling effort that leads to the addition or deletion of conditioning variables or dependencies in a decision model. For example, a decision maker may discover that an expensive telephone-based weather service gives extremely accurate weather forecasts, and wish to include the results of a query to the service in his decision analysis.

These classes of refinement represent distinct dimensions of effort to enhance a decision model. In the next sections, we will develop equations that describe the expected value of continuing to refine a model for each dimension of refinement.

## 2  Expected Values of Decision-Model Refinement

Let us now formalize measures of the expected value of refinement (EVR)[1]. For any dimension of EVR, we seek to characterize our current state of uncertainty about the outcome of an expenditure of effort to refine a decision model. Experienced decision analysts often have strong intuitions about the expected benefits of refining a decision model in different ways. This knowledge is based on expertise, and is conditioned on key observables about the history and state of the modeling process. Assume that we assess and represent such knowledge in terms of probability distributions over the value of the best decision available following model refinement, conditioned on key modeling contexts.

To compute the EVR, we first determine the expected value associated with the set of possible models we create after refinement. We sum together the expected utility of the best decision recommended by each possible revised model, weighted by the likelihood of each model. Finally, we subtract this revised expected value from the expected value of the decision recommended by the unrefined model.

### 2.1  General Analysis

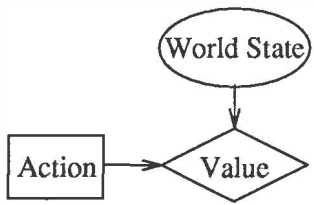

Figure 1: A basic decision model

Consider the simple decision problem with a single state variable $X$ and a single decision variable $A$, as shown in Figure 1. In the party problem, $X$ represents the weather and $A$ represents the decision on party location. The expected value of taking action $a_k$, given background information $\xi$, is

$$E[v|a_k, \xi] = \sum_i p(x_i|\xi) v(a_k, x_i). \quad (1)$$

The expected value of the decision offered by this decision model is

$$E[v|\xi] = \max_k \sum_i p(x_i|\xi) v(a_k, x_i). \quad (2)$$

Suppose the decision model can be refined via one of several refinement procedures $R$. In general, $R$ can be parameterized by amount of effort (e.g., as characterized by time) expended on the refinement. We shall simplify our presentation by initially overlooking such a parameterization. Note that $\xi$ represents the state prior to any refinement consideration; $R$ represents information about the refinement prior to actual refinement. Let $R(\xi)$ denote the state of information after a refinement requiring some prespecified effort. Let $\mu_k$ denote the expected utility that will be obtained for action $a_k$. Before the refinement is carried out, the values of $\mu_k$ are unknown. However, we can assess a probability distribution over each of the values, given information about $R$ and $\xi$. We denote this distribution as $p(\mu_k|R, \xi)$. The expected utility given refinement $R$ is

$$E[v|R(\xi)] = \int_{\mu_1 \cdots \mu_m} p(\mu_1, \ldots, \mu_m|R, \xi)[\max_k \mu_k]. \quad (3)$$

If we cease model-refinement activity, we commit to an action in the world based on all information available—including $p(\mu_k|R, \xi)$. The expected utility without refinement is

$$E[v|R, \xi] = \max_k \int_{\mu_k} \mu_k \, p(\mu_k|R, \xi). \quad (4)$$

The EVR is

$$\text{EVR}(R) = E[v|R(\xi)] - E[v|R, \xi]. \quad (5)$$

In practice, the values $\mu_k$ and distributions $p(\mu_k|R, \xi)$ are dependent on the specific type of refinement and the amount of effort allocated. We shall now describe specific properties of the three types of model refinement and give examples of the detailed analysis of computing the EVR for each. In each case, we shall show how each of the analyses is related to the general formulation captured in Equation (5).

### 2.2  Expected Value of Quantitative Refinement

We start with a consideration of the value of efforts to refine quantitative measures of likelihoods and preferences.

---

[1] We shall use the EVR to refer generally to the expected value of refinement, but shall use more specific terms to refer to alternate classes of refinement.

### 2.2.1 Uncertainty Refinement

Consider the quantitative refinement on the state variable $X$ of the party-location problem. What is the value of "extending the conversation" through expending effort to refine the probability distribution $p(X|\xi)$ with additional assessment. Let us first consider the general case where the distribution $p(X|\xi)$ is continuous. Assume that a continuous distribution is characterized or approximated by a named distribution and a parameter or a vector of parameters. Specifically, assume a functional form $f$ for the probability density function, such that for every reasonable distribution $p(X|\xi)$, there exists a parameter or a set of parameters $\beta$, so that the the numerical approximation $p(X|\xi) \approx f_\beta(X)$ is within satisfactory limits. Before the assessment is carried out, we cannot be certain about the outcome distribution; however, its outcome might be described by a distribution of the form $p(\beta|R,\xi)$, which represents the decision maker's uncertainty about the primary distribution parameter $\beta$. The expected value of the refinement is

$$E[v|R(\xi)] = \int_\beta p(\beta|R,\xi)[\max_k \int_x f_\beta(x)v(a_k,x)] \quad (6)$$

The expected value without performing the quantitative refinement but taking account of knowledge an agent has about the *potential* outcome of refinement procedure $R$ is

$$\begin{aligned} E[v|R,\xi] &= \max_k \int_x \int_\beta f_\beta(x) p(\beta|R,\xi) v(a_k,x) \\ &= \max_k \int_x \hat{p}(x|R,\xi) v(a_k,x) \quad (7) \end{aligned}$$

where

$$\hat{p}(X|R,\xi) = \int_\beta f_\beta(x) p(\beta|R,\xi)$$

is the *operative* distribution for the *authentic* distribution $p(X|\xi)$ (Tani, 1978; Logan, 1985).

The operative distribution is the distribution which the decision maker should use if no further assessment is performed. Let $\hat{\beta}$ be the parameter that best approximates the operative distribution $\hat{p}(X|\xi)$, i.e., the numerical approximation $\hat{p}(X|\xi) \approx f_{\hat{\beta}}(X)$ is within satisfactory limits. This is different from $\bar{\beta} = \int_\beta \beta\, p(\beta|R,\xi)$ which denotes the mean of the secondary distribution.

The expected value of quantitative refinement on the uncertainty on $X$ with respect to assessment procedure $R$, denoted $\text{EVR}^{QU}(R)$ is the difference between (6) and (7).

Let us consider the case where the state variable $X$ is discrete with two states $\{x_1, x_2\}$. We are interested in the value of improving the probabilities assessed for $p(x_1|\xi)$ and $p(x_2|\xi)$. We denote the assessed values of $p(x_1|\xi)$ and $p(x_2|\xi)$ by $\pi$ and $1-\pi$, respectively. The parameter which describes the primary distribution over $X$ is $\beta = \pi$, and we have $f_\pi(x_1) = \pi$ and $f_\pi(x_2) = 1-\pi$. Hence $f$ is linear in $\pi$ and therefore $\hat{\pi} = \bar{\pi}$. The expected value given that quantitative refinement is performed is, $E[v|R(\xi)]$

$$= \int_\pi p(\pi|R,\xi) \max_k [\pi v(a_k,x_1) + (1-\pi) v(a_k,x_2)]. \quad (8)$$

The expected value without the refinement but with knowledge about the potential performance of $R$ is

$$E[v|R,\xi] = \max_k [\bar{\pi} v(a_k,x_1) + (1-\bar{\pi}) v(a_k,x_2)]. \quad (9)$$

The above analysis can be extended to the general case where the state variable $X$ has $n$ possible states. In this case, $\beta$ consists of $n-1$ parameters $(\pi_1, \ldots, \pi_{n-1})$. Our analysis of $\text{EVR}^{QU}(R)$ can be related to the general formulation in Equation (5) by defining the variable

$$\mu_k = \pi v(a_k,x_1) + (1-\pi) v(a_k,x_2) \quad (10)$$

for each action $a_k \in A$. The distributions $p(\mu_k|R,\xi)$ can be derived from $p(\pi|R,\xi)$.

We shall illustrate the concept of quantitative refinement with a example drawn from the party problem. Consider the problem of selecting a location for the party given uncertainty about the weather. Let the alternatives for the location be "Outdoor" ($a_1$) and "Indoor" ($a_2$), and let the weather conditions be "Rain" ($x_1$) or "Sunny" ($a_2$). Let $\pi$ denote the probability that it will rain. The utility values are,

|         | Rain ($\pi$) | Sunny ($1-\pi$) |
|---------|--------------|------------------|
| Outdoor | 0.00         | 1.00             |
| Indoor  | 0.67         | 0.57             |

The optimal locations as a function of $\pi$ are

$$a^*(\pi) = \begin{cases} \text{Outdoor} & \text{if } \pi \leq 0.38 \\ \text{Indoor} & \text{if } \pi > 0.38 \end{cases}$$

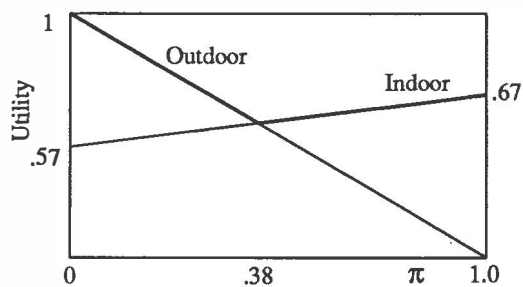

Figure 2: The optimal party location as a function of the probability of rain ($\pi$)

Let us suppose that the current uncertainty about $\pi$ can be described by a probability distribution whose mean is 0.4. In this case, the optimal decision, without further assessment, is to hold the party indoors, with an expected utility of 0.61. However, a more accurate assessment of the value of $\pi$ might change the optimal decision resulting in a potentially higher utility. With refinement,

$$E[v|R(\xi)] = \int_\pi p(\pi|R,\xi)\nu(a^*(\pi),\pi)$$

where

$$\nu(a^*(\pi),\pi) = \begin{cases} 1-\pi & \text{if } \pi \leq 0.38 \\ 0.57 + 0.1\pi & \text{if } \pi > 0.38 \end{cases}$$

Consider the case where $\pi$ is uniformly distributed between the interval [0.3,0.5]. The expected value of refinement, $\text{EVR}^{QU}(R)$, is then,

$$\begin{aligned}
&= \int_{.30}^{.38} 5(1-\pi)d\pi + \int_{.38}^{.50} 5(0.57+0.1\pi)d\pi - 0.61 \\
&= 5[\pi - 0.5\pi^2]_{.3}^{.38} + 5[0.57\pi + 0.05\pi^2]_{.38}^{.5} - 0.61 \\
&= 0.6324 - 0.610 = 0.0224
\end{aligned}$$

Notice that the above analysis was performed in the $\pi$-domain. An alternative analysis and perspective which will produce equivalent results can be performed in the $\mu$-domains. This is done by a change of variables from $\pi$ to $\mu_1$ and $\mu_2$ via Equation (10). The resulting analysis would have to be displayed as a two-dimensional graph.

### 2.2.2 Preference Refinement

Let us now consider the expected value of quantitative refinement of preference $\text{EVR}^{QP}(R)$. We seek to improve the values of $v(a_k, x_i)$ for each $k$ and $i$. Let $\phi_{ki}$ denote the value that will be assessed, given that the refinement is carried out. Let $p(\phi_{ki}|R,\xi)$ denote the uncertainty over the assessment for each $v(a_k, x_i)$. The expected value, given that the quantitative refinement on preference is carried out, is $E[v|R(\xi)]$

$$= \int_{\phi_{11}\cdots\phi_{mn}} p(\phi_{11},\ldots,\phi_{mn}|R,\xi)[\max_k \sum_i p(x_i|\xi)\phi_{ki}]. \tag{11}$$

The expected value without quantitative refinement on preference but with knowledge about the performance of $R$ is

$$E[v|R,\xi] = \max_k \sum_i p(x_i|\xi)\bar{\phi}_{ki} \tag{12}$$

where

$$\bar{\phi}_{ki} = \int_{\phi_{ki}} \phi_{ki}\, p(\phi_{ki}|\xi)$$

is the operative utility value for $v(a_k, a_{ki})$. The $\text{EVR}^{QP}(R)$ is the difference between (11) and (12). This analysis can be related to the general formulation in Equation (5) by defining the variable

$$\mu_k = \sum_i p(x_i|\xi)\phi_{ki} \tag{13}$$

for each action $a_k \in A$. The distributions $p(\mu_k|R,\xi)$ can be derived from the distributions $p(\phi_{ki}|R,\xi)$.

Let us again use the party problem to illustrate the value of refining preferences. Since we can fix the utility for the worst outcome (outdoor and rain) at zero, and the utility for the best outcome (outdoors and sunny) at one, we need only to consider the uncertainty over further assessment of the values $\phi_{21}$ (indoor and rain), and $\phi_{22}$ (indoor and sunny). Let the uncertainty over these values be:

$$\begin{aligned}
\phi_{21} &= U[0.62, 0.72] \\
\phi_{22} &= U[0.52, 0.62]
\end{aligned}$$

The operative values for the preference values are,

|         | Rain (.4) | Sunny (.6) | eu   |
|---------|-----------|------------|------|
| Outdoor | 0.00      | 1.00       | 0.60 |
| Indoor  | 0.67      | 0.57       | 0.61 |

The default choice without any further assessment is to hold the party indoors with an expected utility of 0.61.

$$\begin{aligned}
\mu_1 &= 0.60 \\
\mu_2 &= 0.4\phi_{21} + 0.6\phi_{22}
\end{aligned}$$

In the example, there is no uncertainty over $\mu_1$. The utility, $\mu_2$, displayed in Figure 3, is a linear sum of two uniformly distributed variables, with a triangular distribution $p(\mu_2|R,\xi)$,

$$= \begin{cases} 400(\mu_2 - .56) & \text{if } .56 \leq \mu_2 \leq .61 \\ 20 - 400(\mu_2 - .61) & \text{if } .61 \leq \mu_2 \leq .66 \\ 0 & \text{otherwise} \end{cases}$$

and an expected value of 0.61.

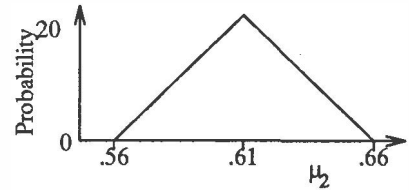

Figure 3: The pdf for $\mu_2$

Figure (4) shows the optimal value $\mu^* = \max_k \mu_k$ as function of $\mu_2$, where $\mu_1$ is fixed at 0.60. The $\text{EVR}^{QP}(R)$ is

$$\begin{aligned}
&= \int_{.56}^{.60} 400(\mu_2 - 0.56)(0.6)d\mu_2 + \\
&\quad \int_{.60}^{.61} [20 - 400(\mu_2 - 0.61)](0.6)d\mu_2 + \\
&\quad \int_{.61}^{.66} [20 - 400(\mu_2 - 0.61)]\mu_2 d\mu_2 - 0.61 \\
&= 0.63733 - 0.610 = 0.02733
\end{aligned}$$

### 2.3 Expected Value of Conceptual Refinement

We shall now explore measures of the value of *conceptual* refinement: (1) the value of refinement of the definitions of state variables, $\text{EVR}^{CS}(R)$ and (2) the refinement of definitions of actions, $\text{EVR}^{CA}(R)$.

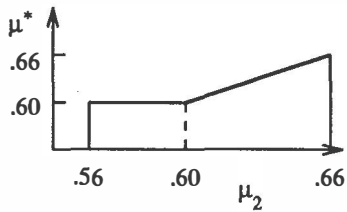

Figure 4: The optimal value $\mu^*$ as a function of $\mu_2$

### 2.3.1 State-Variable Refinement

Assume that our current decision model has a state variable $X = \{x_1, x_2\}$ and decisions $A = \{a_1, a_2\}$. Now, let us consider the value of refining the state $x_1$ into $x_{11}$ and $x_{12}$, such that the resulting state variable is $X' = \{x_{11}, x_{12}, x_2\}$. We further assume that the probability of the refined states $p(x_{11}|x_1, \xi)$ and $p(x_{12}|x_1, \xi)$ are known. As a result of the refinement, we need to assess the utilities $v(a_k, x_{1j})$ for $k = 1, 2$ and $j = 1, 2$. Before these assessments are carried out, the values $v(a_k, x_{1j})$ are unknown. Let $\phi_{kj}$ represent the utilities $v(a_k, x_{1j})$ that will be assessed if the assessment is performed. In addition, we assume that the decision maker is able to assess a set of probability distributions $p(\phi_{kj}|R, \xi)$, $k = 1, 2$ and $j = 1, 2$ over these utilities.

To assess the probabilities over the utilities, a possible conversation between the analyst and the decision maker might be as follows:

> *In our previous conversation, you assigned a utility u for outcomes at your point of indifference between an outcome and a lottery with probability u for the best prospect and probability $1-u$ for the worst prospect. Suppose I were to ask you to assess the utility of each of the refined outcomes. As we do not have an unlimited amount of time to assess these utilities, please give us an estimate now of the probabilities describing the utility values assessed if you were to have enough time to thoroughly reflect on your preferences and knowledge about the outcomes.[2]*

The expected value resulting from the conceptual refinement of $X$ to $X'$ is $E[v|R(\xi)]$

$$= \int_{\phi_{11}\phi_{12}\phi_{21}\phi_{22}} p(\phi_{11}, \phi_{12}, \phi_{21}, \phi_{22}|R, \xi)$$
$$\max_{k=1,2}[p(x_{11}|\xi)\phi_{k1} + p(x_{12}|\xi)\phi_{k2} + \quad (14)$$
$$p(x_2|\xi)v(a_k, x_2)].$$

---
[2]We could also perform this assessment in terms of the utilities that would be assessed after some predefined amount of time for reflection.

The expected value given that the refinement is not carried out, but with knowledge about the performance of $R$ is $E[v|R, \xi]$

$$= \max_{k=1,2}[p(x_{11}|\xi)\bar{\phi}_{k1} + p(x_{12}|\xi)\bar{\phi}_{k2} +$$
$$p(x_2|\xi)v(a_k, x_2)] \quad (15)$$

where $\bar{\phi}_{kj} = \int_{\phi_{kj}} \phi_{kj}\, p(\phi_{kj}|\xi)$, ($k = 1, 2$ and $j = 1, 2$) is the operative value to be used when no refinement is carried out. The $\text{EVR}^{CS}(R)$ for refining state variable $X$ to $X'$ is just the difference between (14) and (15). As before, we can simplify this analysis and relate it to the general formulation of Equation (5) by defining the variable $\mu_k$

$$= p(x_{11}|\xi)\phi_{k1} + p(x_{12}|\xi)\phi_{k2} + p(x_2|\xi)v(a_k, x_2), \quad (16)$$

for action $a_k$, $k = 1, 2$ and deriving the distributions $p(\mu_k|R, \xi)$ from the distributions $p(\phi_{kj}|R, \xi)$.

To illustrate conceptual refinement, consider the expansion of the state of "Rain" into "Downpour" and "Drizzle". Assume that a decision maker's assessment of his uncertainty over the values of $\phi_{12}$ (outdoor and drizzle), $\phi_{21}$ (indoor and downpour), and $\phi_{22}$ (indoor and drizzle) are as follows:

$$p(\phi_{12}|R, \xi) = U[0.05, 0.15]$$
$$p(\phi_{21}|R, \xi) = U[0.67, 0.77]$$
$$p(\phi_{22}|R, \xi) = U[0.57, 0.67]$$

The operative utilities are as follows:

|  | Rain (.4) | | Sunny (.6) | EV |
|---|---|---|---|---|
|  | Downpour (.2) | Drizzle (.2) |  |  |
| Outdoor | 0.00 | 0.10 | 1.00 | 0.62 |
| Indoor | 0.72 | 0.62 | 0.57 | 0.61 |

Without refinement, the expected utility of holding the party outdoors is 0.62 and the utility of having the party indoors is 0.61. Since the two expected values are very close, further refinement might lead to a better discrimination between the two choices. In lieu of additional refinement, the default decision is to have the party outdoors. Based on the distributions over $\phi_{ki}$, we define

$$\mu_1 = 0.2\phi_{12} + 0.6$$
$$\mu_2 = 0.2\phi_{21} + 0.2\phi_{22} + 0.342$$

where $\mu_1$ is uniformly distributed between 0.61 and 0.63, i.e. $p(\mu_1|R, \xi) = U[0.61, 0.63]$, while $\mu_2$ has a triangular distribution $p(\mu_2|R, \xi)$ (depicted in Figure 5),

$$= \begin{cases} 2500(\mu_2 - .59) & \text{if } .59 \leq \mu_2 \leq .61 \\ 50 - 2500(\mu_2 - .61) & \text{if } .61 \leq \mu_2 \leq .63 \\ 0 & \text{otherwise.} \end{cases}$$

Figure (6) shows the region over possible values of $\mu_1$ and $\mu_2$. The $\text{EVR}^{CS}(R)$ is

$$= \int_{.61}^{.63} 50 \left[ \int_{.59}^{.61} 2500(\mu_2 - 0.59)\mu_1 d\mu_2 + \right.$$

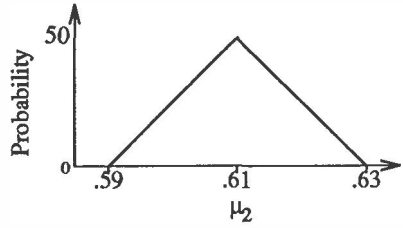

Figure 5: The pdf for $\mu_2$

$$\int_{.61}^{\mu_1} [50 - 2500(\mu_2 - 0.61)]\mu_1 d\mu_2 +$$
$$\left.\int_{\mu_1}^{.63} [50 - 2500(\mu_2 - 0.61)]\mu_2 d\mu_2 \right] d\mu_1$$
$$-0.620 = 0.9208 - 0.620 = 0.3008$$

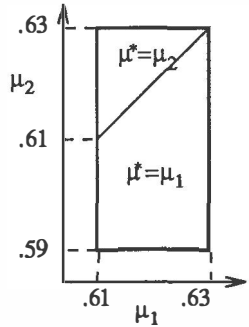

Figure 6: The region of values for $\mu_1$ and $\mu_2$ where $\mu^* = \max_k \mu_k$

### 2.3.2 Action Refinement

Similar to extending the conversation about the definition of states, the set of decision alternatives may be increased with continuing modeling effort. Consider the conceptual refinement of action $A = \{a_1, a_2\}$ by the addition of action $a_3$. Let $A' = \{a_1, a_2, a_3\}$. Unlike state variable refinement, the set of refined actions need not be mutually exclusive. Indeed, they need not even be mutually exhaustive as some alternatives can be ruled out immediately, based on common sense knowledge or dominance relationships (Wellman, 1988). As the result of action refinement we need to assess the utilities $v(a_3, x_i)$ for all $x_i \in X$. Let $\phi_i$ denotes the utility $v(a_3, x_i)$ for each $i$, and let $p(\phi_i|R, \xi)$ be the uncertainty over each assessment. The expected value offered by the refined model is $E[v|R(\xi)]$

$$= \int_{\phi_1 \cdots \phi_n} p(\phi_1, \ldots, \phi_n|R,\xi)[\max_{k=1,2,3}\sum_i p(x_i|\xi)u_{ki}]$$
(17)

where
$$u_{ki} = \begin{cases} v(a_k, x_i) & \text{if } k = 1, 2 \\ \phi_i & \text{if } k = 3. \end{cases}$$

The expected value without the conceptual refinement on action is,

$$E[v|R,\xi] = \max_{k=1,2,3} \sum_i p(x_i|\xi)\bar{u}_{ki} \qquad (18)$$

where
$$\bar{u}_{ki} = \begin{cases} u_{ki} = v(a_k, x_i) & \text{if } k = 1, 2 \\ \bar{\phi}_i & \text{if } k = 3. \end{cases}$$

The $\text{EVR}^{CA}(R)$ for refining action $A$ to $A'$ is then the difference between (17) and (18). We can relate these results to the general formulation of Equation (5) by defining the variable

$$\mu_k = \sum_i p(x_i|\xi)u_{ki} \qquad (19)$$

for each action $a_k$. Note that $\mu_1$ and $\mu_2$ are deterministic, while the probability distribution $p(\mu_3|R,\xi)$ can be derived from the distributions $p(\phi_i|R,\xi)$.

Let us consider the refinement of the example problem with the addition of a third action which—to hold the party on the porch ($a_3$). To complete the refinement, we must assess the utility values $\phi_1$ (porch and downpour), $\phi_2$ (porch and drizzle), and $\phi_3$ (porch and sunny). For simplicity, we will assume that the decision maker is certain about the value of $\phi_3$, which is 0.81. His uncertainty over $\phi_1$ and $\phi_2$ are

$$\phi_1 = U[0.17, 0.27]$$
$$\phi_2 = U[0.37, 0.47]$$

The operative utilities are as follows:

|  | Rain (.4) | | | |
|---|---|---|---|---|
|  | Down-pour (.2) | Drizzle (.2) | Sunny (.6) | EV |
| Outdoor | 0 | .10 | 1 | .620 |
| Indoor | .72 | .62 | .57 | .610 |
| Porch | .22 | .42 | .81 | .614 |

The optimal action without further refinement is to hold the party outdoors, with an expected utility of 0.62.

$$\mu_1 = 0.62$$
$$\mu_2 = 0.61$$
$$\mu_3 = 0.2\phi_1 + 0.2\phi_2 + 0.6\phi_3$$

There is no uncertainty on $\mu_1$ and $\mu_2$. However, as displayed in Figure 7, $\mu_3$ is a linear sum of two uniformly distributed variable and has a triangular distribution of the form, $p(\mu_3|R,\xi)$,

$$= \begin{cases} 400(\mu_3 - .564) & \text{if } .564 \leq \mu_3 \leq .614 \\ 20 - 400(\mu_3 - .614) & \text{if } .614 \leq \mu_2 \leq .664 \\ 0 & \text{otherwise} \end{cases}$$

and has an expected value of 0.614.

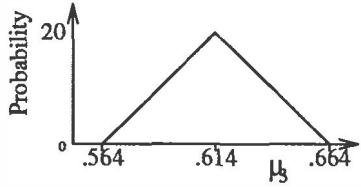

Figure 7: The pdf for $\mu_2$

Figure (8) shows the optimal value $\mu^* = \max_k \mu_k$ as a function of $\mu_3$. The expected value of conceptual refinement via addition of the third alternative is

$$= \int_{.564}^{.614} 400(\mu_3 - 0.564)(0.62)d\mu_3 +$$
$$\int_{.614}^{.620} [20 - 400(\mu_3 - 0.614)](0.62)d\mu_3 +$$
$$\int_{.620}^{.664} [20 - 400(\mu_3 - 0.614)]\mu_3 d\mu_3 - 0.62$$
$$= 0.62568 - 0.62 = 0.00568$$

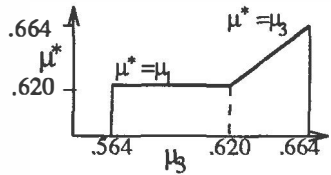

Figure 8: The optimal value $\mu^*$ as a function of $\mu_3$

### 2.4 Expected Value of Structural Refinement

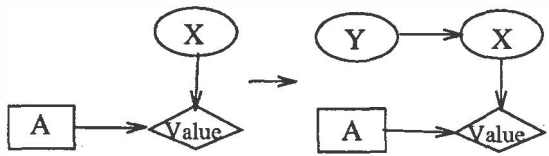

Figure 9: Structural refinement on node $X$

Finally, we consider the value of *structural* refinement, $EVR^S(R)$, the value of increasing the number of conditioning variables. Figure 9 depicts an extension of conversation based on structural refinement of the state variable $X$ of our simple decision problem by the addition of $Y$ as a conditioning event for $X$. For example, in the party problem, we may identify "wind speed" as a conditioning variable on the forthcoming weather. We are interested in analyzing the additional value that is gained by the addition of $Y$ as a conditioning variable for $X$. This structural refinement requires the assessment of the probability distributions $p(Y|R(\xi))$ and $p(X|Y, R(\xi))$. As before, we assume a functional form $f$ where, for every reasonable distribution for $p(Y|R(\xi))$, there exists a parameter $\beta_Y$, such that the numerical approximation $p(Y|R(\xi)) \approx f_{\beta_Y}(Y)$ is within satisfactory limits. We let $\beta_{X|Y}$ represent the parameter for the distribution $p(X|Y, R(\xi))$. Let the distributions $p(\beta_Y|R, \xi)$ and $p(\beta_{Y|X}|R, \xi)$ represent the decision maker's uncertainty about the parameters $\beta_Y$ and $\beta_{X|Y}$, respectively. The expected value that results from the structural refinement via the addition of $Y$ as a conditioning variable for $X$, is $E[v|R(\xi)]$

$$= \int_{\beta_Y} p(\beta_Y|R,\xi) \int_{\beta_{Y|X}} p(\beta_{Y|X}|R,\xi)$$
$$\left[\max_k \int_y f_{\beta_Y}(y) \int_x f_{\beta_{X|Y}}(x)v(a_k,x)\right]. \quad (20)$$

The expected value without structural refinement is

$$E[v|R,\xi] = \max_k \int_y f_{\hat{\beta}_Y}(y) \int_x f_{\hat{\beta}_{X|Y}}(x)v(a_k,x), \quad (21)$$

where $\hat{\beta}_Y$ and $\hat{\beta}_{X|Y}$ are the parameters for the operative distributions

$$\hat{p}(Y|R,\xi) = \int_{\beta_Y} f_{\beta_Y}(Y)p(\beta_Y|R,\xi) \approx f_{\hat{\beta}_Y}(Y)$$

and

$$\hat{p}(X|Y,R,\xi) = \int_{\beta_{X|Y}} f_{\beta_Y}(X)p(\beta_{X|Y}|R,\xi) \approx f_{\hat{\beta}_{X|Y}}(X)$$

respectively. The $EVR^S(R)$ for the variable $X$, with respect to adding a new conditioning event $Y$, is just the difference between (20) and (21). The case where $X$ and $Y$ are discrete variables is treated in (Poh & Horvitz, 1992).

A special form of structural refinement is the familiar expected value of information ($EVI$). Within the influence diagram representation, we can view the observation of evidence as the addition of arcs between chance nodes and decisions. We describe the relationship of EVI and other dimensions of model refinement in (Poh & Horvitz, 1992).

## 3 Control of Refinement

Measures of EVR, computed from a knowledge base of probabilistic expertise about the progress of model refinement, hold promise for providing guidance in controlling decision modeling in consultation settings, as well as within automated decision systems. In this section, consider control techniques for making decisions about the refinement of decision models.

### 3.1 Net Expected Value of Refinement

So far, we have considered only the *value* of alternative forms of effort to expending effort to refine a model. To consider the use of EVR measures, we must balance the expected benefits of model refinement with (1) the cost of the assessment effort, and (2) the increased computational cost of solving more refined, and potentially more complex, decision models. We define the the net expected value of refinement, NEVR, as the difference between the EVR and the cost of making a refinement and increase in the cost of solving the refined model. That is NEVR$(R,t)$

$$= \text{EVR}[(R(t)),\xi] - C_a(t_a) - C_c(\Delta(t_c)) \qquad (22)$$

where $R(t)$ is a refinement parameterized by the time expended on a particular refinement procedure, $C_a$ is a function converting assessment time, $t_a$ to cost, and $C_c$ is a function converting changes in the expected computational time, required to solve the decision problem, $\Delta(t_c)$, to cost. In offline, consultation settings, we can typically assume that changes in computational costs, associated with the solving decision models of increasingly complexity, are insignificant compared with the costs of assessment. We can introduce uncertainty into the costs functions with ease.

### 3.2 Decisions about Alternative Refinements

Let us assume that we wish to identify the best refinement procedure to extend a decision model. For now, let us assume that we have deterministic knowledge about the cost of refinements. We shall assume that the cost is a deterministic function of time[3] and that computational changes with refinement are insignificant.

We can control model building with a strategic optimization (Horvitz, 1990) that seeks to identify the best refinement procedure and the amount of effort to allocate to that procedure, i.e.,

$$\arg\max_{R,t} \text{EVR}[(R(t)),\xi] - C(t) \qquad (23)$$

Given appropriate knowledge about decision model refinement, we solve such a maximization problem by computing the ideal amount of effort to expend for each available refinement methodology, choose the procedure $R^*$ with the greatest NEVR, and apply it for the ideal amount of time, $t^*$ computed from the maximization. We halt refinement when all procedures have NEVR$(R,t) < 0$ for all times $t$.

However, we need not be limited to considering single procedures. In a more general analysis, we allow for the interleaving of arbitrary sequences of refinement procedures, where each refinement procedure can be allocated an arbitrary amount of effort, and to consider sequences of refinements with the greatest expected value. As any refinement changes a model, and, thus, changes the value of refinement for future modeling efforts, the identification of a theoretically optimal sequence requires a combinatorial search through all possibilities. Let us consider several approximations to such an exhaustive search.

---

[3]In practice, a decision consultant may wish to consider such multiattribute cost models as the cost in time, dollars, and frustration associated with pursuit of different kinds of assessments and refinements.

A practical approach to dodging the combinatorial control problem is to consider predefined quantities of effort, and to employ a myopic or *greedy* EVR control procedure. With a greedy assumption, we simplify our analysis of control strategies by making the typically invalid assumption that we will halt modeling, solve the decision model, and take an action following a *single expenditure* of modeling effort. We can further simplify such a myopic analysis by assuming a predefined, constant amount of effort to employ in NEVR analyses. We compute the EVR$(R(T))$ for all available refinement procedures $R$, where $T$ is some constant amount of time, or a quantity of time $T_R = T(R)$, a constant amount of time keyed to specific procedures. At each cycle, we compute the NEVR for all procedures, and implement the refinement procedure with the greatest NEVR. We iteratively repeat this greedy analysis until the cost of all procedures is greater than the benefit, at which time we solve the decision problem and take the recommended action. Figure (10) shows a fragment of the graph of possible model refinement steps.

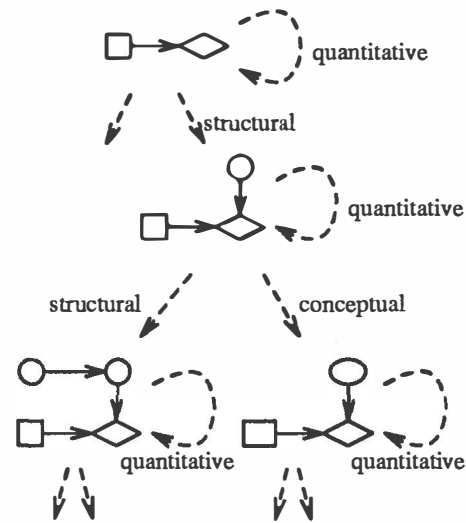

Figure 10: Greedy control of model refinement with iterative application of NEVR analyses

We can relax the myopia of the greedy analysis by allowing varying amounts of lookahead. For example, we can consider the NEVR of two refinement steps. Such lookahead can be invoked when single steps yield a negative NEVR for all refinement methods. We can also make use of theoretical dominance results. For example, we have shown in a more comprehensive paper that the expected value of perfect information (EVPI) is the upper bound on the value of any structural refinement (Poh & Horvitz, 1992).

## 4  Discussion and Related Work

The value of the EVR methods hinges on the availability of probability distributions that describe the outcomes of extending models in different ways. We suspect that expert analysts rely on such probabilistic modeling metaknowledge, and that relatively stable probability distributions can be assessed for prototypical contexts and states of model completeness. We do not necessarily have to rely on assessing an expert decision analyst's probability distributions about alternative outcomes of modeling. In an automated decision support setting, we can collect statistics about modeling and modeling outcomes. Such data collection can be especially useful for the application of EVR-based control strategies to automated reasoning systems that construct models dynamically (Breese, 1987; Goldman & Breese, 1992).

We are not the first to explore the value of modeling in decision analysis. The value of modeling was first addressed by Watson and Brown (1978) and Nickerson and Boyd (1980). The notion of reasoning about the value of probability assessment with an explicit consideration of how second-order distributions change with assessment effort has been explored rigously by Logan (1985). Chang and Fung (1990) have considered the problem of dynamically refining and coarsening of state variables in Bayesian networks. They specified a set of constraints that must be satisfied to ensure that the coarsening and weakening operations do not affect variables that are not involved. In particular, the joint distribution of the Markov blanket excluding the state variable itself must be preserved. However, the value and cost of performing such operations were not addressed. The form of refinement that we refer to as structural refinement has also been examined by Heckerman and Jimison (1987) in their work on attention focusing in knowledge acquisition. Finally, related work on control of reasoning and rational decision making under resource constraints, using analyses of the expected value of computation and considering decisions about the use of alternative strategies and allocations of effort, has been explored by Horvitz (1987, 1990) and Russell and Wefald (1989).

## 5  Summary and Conclusions

We introduced and distinguished the expected value of quantitative, conceptual, and structural refinement of decision models. We believe that the analyses of the value of model refinement hold promise for controlling the attention of decisions makers, and of automated reasoning systems, on the best means of extending a decision model. Such methods can also be employed to determine when it is best to halt refinement procedures and instead to solve a decision model to identify a best action. We look forward to assessing expert knowledge about the value of decision-model refinement and testing these ideas in real decision analyses. We are striving to automate the assessment of knowledge about model refinement, as well as the iterative cycle of EVR computation. We are implementing key ideas described in this paper within the IDEAL influence diagram environment (Srinivas & Breese, 1990).